%% file: main.tex
\documentclass[letterpaper, 10 pt, conference]{ieeeconf}  %

\usepackage{booktabs}
\usepackage{graphicx}
\usepackage[table,xcdraw]{xcolor}

\IEEEoverridecommandlockouts                              %

\usepackage{graphics} %
\usepackage{epsfig} %
\usepackage{mathptmx} %
\usepackage{times} %
\usepackage{amsmath} %
\usepackage{amssymb}  %
\usepackage{xcolor} %

\usepackage{tikz}
\usepackage{pgfplots}
\usepgfplotslibrary{groupplots}
\usepackage{tikzscale}
\usepackage{adjustbox}
\usepackage{balance}

\input{vars.tex}

\title{\LARGE \bf
\papername{}: Geometry-Aware Refinement for Lane Segment Detection and Topology Reasoning
}

\author{
Danny Abraham$^{1}$,
Nikhil Kamalkumar Advani$^{2}$,
Arun Das$^{2}$,
Nikil Dutt$^{1}$%
\thanks{$^{1}$University of California, Irvine, CA, USA
{\tt\small dannya1@uci.edu, dutt@uci.edu}}%
\thanks{$^{2}$Bosch North America, Sunnyvale, CA, USA
{\tt\small nik.advani16@gmail.com, Arun.Das@us.bosch.com}}%
}

\begin{document}

\maketitle
\thispagestyle{empty}
\pagestyle{empty}

\begin{abstract}

Accurate 3D lane segment detection and topology reasoning are critical for structured online map construction in autonomous driving.
Recent transformer-based approaches formulate this task as query-based set prediction, yet largely inherit decoder designs originally developed for compact object detection.
However, lane segments are continuous polylines embedded in directed graphs, and generic query initialization and unconstrained refinement do not explicitly encode this geometric and relational structure.
We propose \papername{} (Geometry-aware Refinement Transformer), a unified query-based architecture that embeds geometry- and topology-aware inductive biases directly within the transformer decoder.
\papername{} introduces data-driven geometric priors for structured query initialization, bounded coordinate-space refinement for stable polyline deformation, and per-query gated topology propagation to selectively integrate relational context.
On the OpenLane-V2 benchmark, \papername{} achieves state-of-the-art performance with 34.5\% mAP while improving topology consistency over strong transformer baselines,
demonstrating the utility of explicit geometric and relational structure encoding.

\end{abstract}

\section{INTRODUCTION}

Autonomous driving systems require structured and reliable perception of road geometry to enable downstream tasks such as motion planning, trajectory optimization, and safety-critical decision making.
Beyond detecting individual lane markings, modern online mapping systems must recover directed lane centerlines, boundary types, and connectivity relationships that define the navigable road graph.
Modern benchmarks such as OpenLane-V2~\cite{wang2023openlanev2} formalize this problem as unified 3D lane detection and topology reasoning, providing large-scale multi-view datasets with metrics that jointly evaluate geometric fidelity and relational consistency.

Transformer-based architectures inspired by the Detection Transformer (DETR) paradigm~\cite{carion2020end} have become the dominant framework for query-based lane mapping.
These models represent lanes as sets of learnable queries that interact with bird’s-eye-view (BEV) features through cross-attention, with geometric predictions progressively updated across decoder layers.
While these formulation offers strong global context modeling and flexible set prediction, most existing approaches inherit decoder designs originally developed for compact object detection, with limited adaptation to the structural properties of lane geometry.

Unlike bounding boxes, lane segments are elongated polylines with strong geometric continuity and explicit connectivity relationships.
However, most query-based lane mapping approaches inherit decoder designs originally developed for compact object detection.
As a result, query initialization and geometric refinement are typically unconstrained, providing little inductive bias toward the spatial continuity and relational structure that characterize road layouts.
This mismatch can limit the decoder’s ability to produce geometrically consistent lane representations and coherent topology predictions.
In many architectures, topology reasoning is further separated into a parallel prediction head rather than integrated into the feature refinement process itself.
Consequently, geometric estimation and relational reasoning are only weakly coupled during decoding.
Finally, because the learned queries are initialized without explicit geometric structure, individual queries often lack consistent semantic roles.
This makes it difficult to interpret which geometric patterns a given query is responsible for detecting, reducing transparency in how the model constructs the predicted lane graph.

\begin{figure}[t]
	\centering
    \includegraphics[width=\columnwidth]{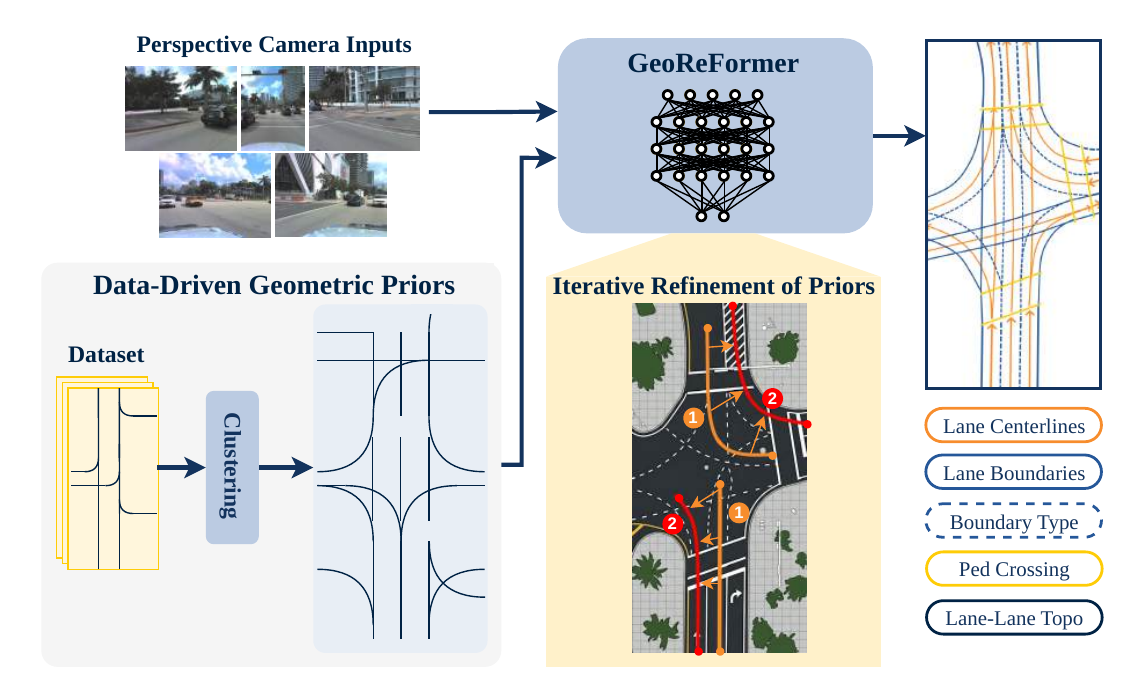}
    \caption{Overview of \textit{\papername{}}, a geometry-aware query-based transformer for joint 3D lane segment detection and topology estimation from multi-view cameras}
	\label{fig:motivation}
\end{figure}

In this work, we introduce \textbf{\papername{}} (\textbf{Geo}metry-aware \textbf{Re}finement Trans\textbf{former}), a unified query-based architecture for joint 3D lane segment detection and topology estimation.
\papername{} is built upon three complementary design principles that explicitly align the transformer decoder with the structure of lane geometry and connectivity.
Fig.~\ref{fig:motivation} shows \papername{}'s three-stage design flow. 
First, we incorporate data-driven geometric priors by clustering normalized lane polylines and using the resulting prototypes to initialize decoder reference points with structured spatial bias.
Second, we reformulate iterative reference updates to operate directly in normalized coordinate space with bounded residual corrections, ensuring geometrically interpretable and controlled polyline refinement across decoder layers.
Third, we integrate topology-aware feature propagation within the decoder by replacing the standard feed-forward network with a topology-gated graph convolution module that enables relational information flow among connected lane queries during feature transformation.

Together, these components embed geometric and relational inductive biases directly into the decoding process.
This design not only improves detection accuracy but also encourages queries to specialize toward consistent geometric roles, enabling interpretable query behavior.

In summary, our contributions are:
\begin{itemize}
    \item \textbf{\papername{}:} A geometry-aware query-based transformer architecture for joint 3D lane detection and topology reasoning.
    \item \textbf{Data-Driven Geometric Prior Initialization:} A clustering-based mechanism that embeds structured spatial bias into decoder reference points.
    \item \textbf{Bounded Coordinate-Space Refinement:} A geometrically interpretable residual update rule that constrains iterative polyline deformation in normalized coordinate space.
    \item \textbf{Topology-Gated Feature Propagation:} A decoder-level graph convolution mechanism that integrates relational reasoning directly into feature refinement.
\end{itemize}

\section{RELATED WORK}

Online HD map construction aims to recover structured, vectorized road geometry and connectivity directly from onboard sensors.
Early approaches predicted rasterized BEV semantics followed by post-processing to extract vector maps~\cite{li2022hdmapnet}.
Recent benchmarks such as OpenLane-V2~\cite{wang2023openlanev2} formalize unified 3D lane detection and topology reasoning, requiring joint prediction of polylines and directed connectivity, extending prior 3D lane perception works~\cite{chen2022persformer} with relational evaluation.

Transformer-based set prediction inspired by DETR~\cite{carion2020end} has become the dominant paradigm for online mapping.
In this formulation, a fixed set of learnable queries interacts with BEV features to predict vectorized lane elements end-to-end~\cite{liu2023vectormapnet,liaomaptr,li2023lanesegnet,yang2025topo2seq}.
Subsequent works enhance topology reasoning using relational propagation or scene-graph modules~\cite{futopopoint,li2023toponet,fu2024topologic,lv2025t2sg}.
Despite architectural variations, most methods inherit generic DETR-style query initialization and logit-space residual updates originally designed for compact object detection, where reference points are learned without explicit geometric structure and geometric refinement and topology reasoning remain loosely coupled.

To improve robustness, several approaches extend query-based decoding to streaming or multi-frame settings by propagating queries and features across time~\cite{yuan2024streammapnet,yang2025topostreamer,wang2024stream}.
Others incorporate structural priors from standard-definition maps or satellite imagery to guide geometry and connectivity prediction~\cite{zhang2024enhancing,ma2024roadpainter,luo2024augmenting,ye2025smart,pham2025coherent}.
While effective, these strategies introduce additional supervision, memory overhead, or reliance on external infrastructure rather than strengthening the decoder’s internal geometric reasoning.

In contrast, relatively limited attention has been given to the refinement mechanism within the transformer decoder itself.
Existing approaches typically adopt unconstrained logit-space updates and treat topology reasoning as a separate head or post-hoc association module.
\papername{} addresses this gap by embedding data-driven geometric priors for structured query initialization, bounded coordinate-space polyline refinement that preserves locality, and per-query gated topology propagation directly within the decoder.

\section{\textit{\papername{}}}

\begin{figure*}[t]
	\centering
    \includegraphics[width=\textwidth]{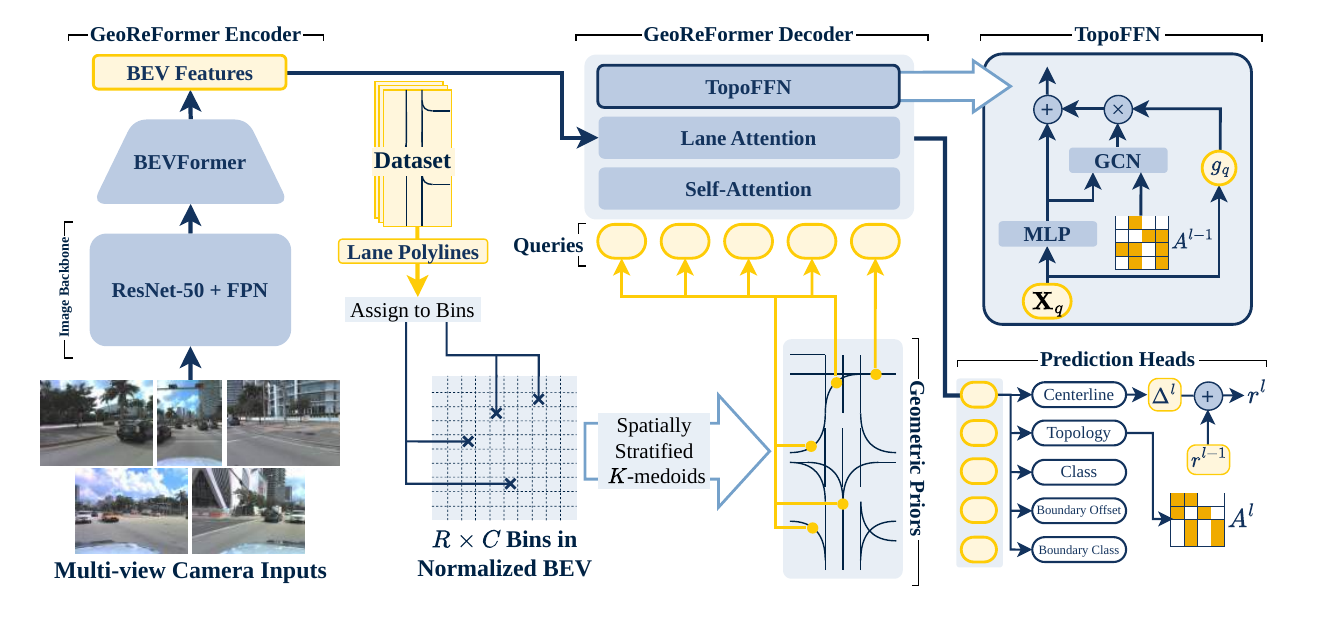}
    \caption{Architecture of \textit{\papername{}}. Multi-view images are lifted to BEV features and decoded from fixed geometric prios obtained via Spatially-Stratified K-Medoids. The decoder integrates bounded polyline refinement and per-query gated topology aggregation (TopoFFN). Prediction heads produce lane geometry and relational topology.}
	\label{fig:architecture}
\end{figure*}

We present \textbf{\papername{}} (\textbf{Geo}metry-aware \textbf{Re}finement Trans\textbf{former}), a query-based detector for joint 3D lane segment prediction and topology estimation from multi-view cameras.
\papername{} builds upon the query-based detection paradigm of~\cite{li2023lanesegnet} and follows a three-stage pipeline: 
(1) multi-view image features are extracted and lifted into a unified 
BEV representation, 
(2) a set of $\mathbf{Q}$ object queries, each representing a candidate lane segment, are iteratively refined through a transformer decoder that attends to the BEV features, and 
(3) per-query prediction heads produce the final 3D polylines, class labels, boundary types, segmentation masks, and pairwise topology.
The overall architecture of \papername{} is illustrated in Fig.~\ref{fig:architecture}.

We 
highlight three innovations in
this pipeline that target the decoder's ability to leverage geometric and topological structure.
First, we replace the standard learned or random query reference points with geometry-aware polyline priors that encode the spatial layout and dominant orientations of lane segments in the BEV (Sec.~\ref{sec:geometric_priors}).
Second, we reformulate the iterative reference point update as a bounded refinement in normalized coordinate space, where each decoder layer applies a $\mathbf{tanh}$-bounded correction that prevents large destabilizing shifts (Sec.~\ref{sec:bounded_refinement}).
Third, we replace the standard feed-forward network (FFN) in each decoder layer with a topology-gated FFN that augments the standard feed-forward transform with a per-query gated graph convolution over the predicted lane adjacency graph (Sec.~\ref{sec:topology_ffn}).
Together, these components define the \papername{} decoder as illustrated in Fig.~\ref{fig:architecture}.

\subsection{Preliminaries}
\label{sec:preliminaries}
We extract multi-scale features from $N=7$ surround-view cameras using a ResNet-50~\cite{he2016deep} backbone with FPN~\cite{lin2017feature}, producing $K=4$ feature levels at channel dimension $D$.
A BEVFormer~\cite{li2024bevformer} encoder lifts these features into a dense BEV representation $\mathbf{B} \in \mathbb{R}^{H \times W \times D}$ via deformable spatial cross-attention along vertical pillars.

Following the DETR~\cite{carion2020end} paradigm, $Q$ learnable queries are decoded through $L$ transformer layers, each performing self-attention among queries, deformable cross-attention to $\mathbf{B}$
via Lane Attention~\cite{li2023lanesegnet}
and a feed-forward transform.
Per-layer heads produce class scores, 3D polylines (centerline + left/right boundary offsets), boundary marking types, BEV segmentation masks, and pairwise lane-lane topology.
In OpenLane-V2 lane segment detection, lane geometry is represented in BEV as a polyline with a fixed number of sampled points.
Concretely, each lane segment is parameterized by $N$ ordered control points, and the decoder predicts these points (or equivalent offsets) for each query.
Training uses Hungarian matching with deep supervision across all decoder layers.

\subsection{Geometric Priors via Spatially-Stratified K-Medoids}
\label{sec:geometric_priors}

We construct the geometric priors of \papername{} using a data-driven procedure termed \textit{Spatially-Stratified K-medoids}, which produces a fixed set of $Q=200$ representative lane polylines that preserve realistic geometry while maintaining balanced spatial coverage over the BEV.

Lane polylines are extracted from the training dataset, each consisting of ten ordered $(x,y)$ coordinates in ego-centric meters.
All polylines are normalized to the unit square corresponding to the BEV extent, and all subsequent binning and clustering operate in this normalized space.

\paragraph{Spatial Stratification}
We partition the normalized BEV into a coarse grid of $R \times C$ bins with $RC < Q$, as shown in Fig.~\ref{fig:architecture}.
In practice, we use a $14 \times 14$ grid (196 bins).
Each polyline is assigned to a bin according to the centroid of its normalized coordinates.
Let $n_k$ denote the number of polylines assigned to bin $k$.

\paragraph{Density-Aware Budget Allocation.}
To account for varying data density, we allocate priors to each occupied bin using a density-aware weighting scheme.
Each bin is assigned a weight
\[
w_k = \max\!\left(1, \sqrt{n_k}\right)
\]
which scales allocation with local data frequency while preventing overrepresentation of dense regions.
The number of priors assigned to bin $k$ is computed as
\[
q_k = \left\lfloor \frac{w_k}{\sum_{\ell} w_{\ell}} \cdot Q \right\rfloor
\]
followed by a largest-remainder correction to ensure $\sum_k q_k = Q$.
Because $RC < Q$, every occupied bin receives at least one prior, guaranteeing spatial support across the BEV.

\paragraph{Within-Bin K-Medoids Selection}
Within each bin $k$, we select $q_k$ representative polylines using K-medoids clustering under a Euclidean distance metric.
K-medoids selects representatives directly from the observed polylines, ensuring that each geometric prior corresponds to a valid, physically realizable lane configuration rather than a synthetic average.

The resulting Spatially-Stratified K-medoids procedure produces a compact set of spatially balanced and geometrically realistic priors.
By combining global BEV coverage, density-aware allocation, and medoid-based selection, the prototypes reflect the empirical distribution of lane geometry without overrepresenting dominant regions.
Importantly, these geometric priors are fixed and non-trainable.
Unlike learned positional reference embeddings used in prior transformer-based mapping models, they remain constant during training and serve as an explicit geometric inductive bias.
This decouples structural initialization from feature learning and provides a stable foundation for the bounded refinement process described next.

\subsection{Bounded Coordinate-Space Polyline Refinement}
\label{sec:bounded_refinement}

In query-based lane mapping models, each decoder layer refines reference polylines through residual prediction. 
These polylines both approximate lane geometry and provide reference points for deformable cross-attention (Lane Attention~\cite{li2023lanesegnet}), guiding feature aggregation from the BEV representation.
Thus, the parameterization of reference updates directly affects both geometric regression and attention sampling.

In standard DETR-style formulations~\cite{zhudeformable, li2023lanesegnet, yang2025topostreamer, yang2025topo2seq}, refinement is performed in logit space.
Given reference points $\mathbf{r}^{l-1}$ at decoder layer $l-1$, the network predicts an unconstrained offset $\Delta^l$, and the updated reference is computed as
\[
\mathbf{r}^{l} = \sigma\!\left( \sigma^{-1}(\mathbf{r}^{l-1}) + \Delta^{l} \right),
\]
where $\sigma$ denotes the sigmoid function.
This parameterization is well suited to single-point bounding box regression, but it does not explicitly account for the geometric structure of multi-point polylines.

Indeed, this formulation introduces two limitations when applied to lane polyline refinement.
First, the offset $\Delta^l$ is unbounded, allowing a single decoder layer to induce arbitrarily large geometric displacements. 
Second, updates occur in logit space, so equal-magnitude offsets produce non-uniform spatial shifts depending on the reference location in $[0,1]$.
As a result, the geometric meaning of a residual update is position-dependent and difficult to interpret.

These limitations are particularly problematic in \papername{}, where reference polylines are initialized from geometric priors (Sec.~\ref{sec:geometric_priors}). 
To preserve the locality encoded by these priors, we perform refinement directly in normalized coordinate space.

Specifically, we represent each polyline reference as $\mathbf{r}^{l-1} \in [0,1]^{N \times 3}$, where $N$ denotes the number of sampled polyline points.
At decoder layer $l$, the regression head predicts a raw correction $\Delta^l \in \mathbb{R}^{Q \times N \times 3}$ from the query embeddings.
We update the reference using
\[
\mathbf{r}^{l} = \mathbf{r}^{l-1} + s \cdot \tanh(\Delta^{l}),
\]
where $s$ is a fixed residual scaling factor.
In practice, we set $s = 0.25$.
The $\tanh$ nonlinearity bounds each coordinate update to $[-s, s]$, ensuring that no layer displaces a polyline point by more than a fixed fraction of the BEV extent.

Operating directly in coordinate space yields two benefits. 
Residual updates have a uniform geometric interpretation: equal-magnitude corrections produce equal displacements in normalized BEV coordinates, and bounded updates encourage incremental, locality-preserving refinement.

Decoder training employs deep supervision with losses applied at each layer, directly penalizing large deviations from ground truth in early iterations. 
Although successive layers can accumulate bounded updates across $L$ iterations, the training objective encourages each prior to match a nearby ground-truth lane rather than traversing large BEV distances. 
This interaction between geometric priors, bounded refinement, and layer-wise supervision promotes consistent alignment between initialization and final prediction.

Overall, bounded coordinate-space refinement complements the structured initialization provided by geometric priors. 
By constraining residual updates to geometrically meaningful increments, \papername{} enables progressive polyline refinement while preserving locality in both regression and attention sampling.

\subsection{Topology-Aware Decoder with Per-Query Gated Aggregation}
\label{sec:topology_ffn}

The preceding sections embed geometric structure into the decoder through data-driven geometric priors and bounded coordinate-space refinement.
We now address the complementary aspect of lane modeling's relational structure.
Lane segments do not exist independently, they form directed connectivity patterns that are essential for topology reasoning and downstream planning.
We therefore incorporate topology directly into the decoder feature updates 
(as shown in Fig.~\ref{fig:architecture})
rather than treating it as a post-hoc classification problem.

Prior works~\cite{fu2024topologic} introduce graph convolutional reasoning over predicted lane relations.
However, these formulations typically replace or serially stack a GCN after the feed-forward network (FFN), causing topology features to dominate all query representations uniformly.
In contrast, \papername{} introduces a residual and selectively gated topology pathway that preserves the base decoder behavior while allowing topology to inform feature refinement only where beneficial.

\paragraph{Topology Construction}
At decoder layer $l$, we construct a soft adjacency matrix $\mathbf{A}^l \in [0,1]^{Q \times Q}$ that encodes directed connectivity between predicted lane segments.
The adjacency blends a learned relational similarity with a geometric proximity term capturing endpoint-to-startpoint alignment.
This follows existing topology modeling practice and serves as the relational signal for feature aggregation.
At initialization, $\mathbf{A}^0 = 0$, ensuring that the decoder reduces to the baseline formulation.

\paragraph{Residual Graph Aggregation}
Given query features $\mathbf{X} \in \mathbb{R}^{Q \times D}$, we compute a topology-aggregated representation using a directed graph convolution:

\[
\text{GCN}(\mathbf{X}, \mathbf{A}) 
= \mathbf{X}\mathbf{W}_s 
+ \mathbf{A}\mathbf{X}\mathbf{W}_f 
+ \mathbf{A}^{\mathsf{T}}\mathbf{X}\mathbf{W}_b,
\]

where $\mathbf{W}_s, \mathbf{W}_f, \mathbf{W}_b \in \mathbb{R}^{D \times D}$ model self, successor, and predecessor interactions.
Unlike prior serial formulations that stack a GCN after the feed-forward network (FFN) and allow topology features to overwrite the base representation, we introduce a residual topology pathway that augments, rather than replaces, the MLP output.

Given query features $\mathbf{X} \in \mathbb{R}^{Q \times D}$, we aim to incorporate relational reasoning without disrupting the geometric refinement behavior of the decoder.
A key observation is that lane queries are heterogeneous: lane segments benefit from relational aggregation, whereas pedestrian crossings are often topologically isolated.
Uniformly injecting topology into all queries can therefore introduce noise.

To enable selective relational reasoning, we introduce a learned per-query scalar gate:

\[
g_q = \sigma(\mathbf{w}^\top \mathbf{X}_q + b), 
\quad g_q \in [0,1],
\]

which modulates the topology contribution independently for each query.
Using this gate, topology is injected through a residual pathway parallel to the MLP branch:

\[
\mathbf{H} = \mathbf{X} 
+ \text{MLP}(\mathbf{X}) 
+ g(\mathbf{X}) \cdot \text{GCN}(\text{MLP}(\mathbf{X}), \mathbf{A}),
\]

where $g(\mathbf{X})$ applies $g_q$ to each query feature.

This additive formulation preserves the base geometric update while allowing relational information to augment features only when beneficial.
The gate is initialized with $\mathbf{w}=0$ and $b=-3$, yielding $g_q \approx 0.05$ at initialization.
This zero-biased design ensures that the decoder initially behaves nearly identically to the baseline formulation and gradually activates topology aggregation during training, following residual modulation principles~\cite{bachlechner2021rezero}.

By combining residual topology injection with per-query gating, \papername{} enables relational reasoning to enhance lane-segment queries while suppressing noisy aggregation for topologically isolated elements.
This selective integration of topology complements the geometric priors and bounded refinement introduced earlier, resulting in a decoder that jointly respects geometric continuity and relational structure.

\section{EVALUATION \& RESULTS}

We evaluate \papername{} on the OpenLane-V2~\cite{wang2023openlanev2} Lane Segment Detection benchmark.
We first compare against recent state-of-the-art methods to assess overall performance, followed by targeted ablations analyzing the contributions of geometric priors, bounded polyline refinement, and the TopoFFN module.
All experiments are conducted under identical training protocols on a single node with four NVIDIA A30 GPUs.

\subsection{Dataset \& Metrics}

We evaluate \papername{} on the Subset A split of the OpenLane-V2 benchmark~\cite{wang2023openlanev2}, which provides large-scale multi-view data for structured 3D lane segment detection with topology annotations.
Each frame contains synchronized surround-view camera images and corresponding bird’s-eye-view (BEV) annotations of lane centerlines, lane boundaries, pedestrian crossings, and lane-to-lane connectivity.
The dataset includes over 27k training frames and approximately 4.8k validation frames.

We focus on the \textit{Lane Segment Detection} task, which evaluates geometric lane prediction and lane-to-lane topology consistency.
Pedestrian crossings are evaluated as a separate class, motivating the topology-gated design that selectively incorporates relational context.

Geometric performance is evaluated using distance-based matching under predefined thresholds, with mean Average Precision (mAP) as the primary metric.
Specifically, $AP_{ls}$ and $AP_{ped}$ measure detection quality for lane segments and pedestrian crossings, respectively, while topology reasoning is assessed using $TOP_{lsls}$, which evaluates predicted lane connectivity.

\subsection{Results on \textit{Lane Segment Detection}}

\input{lanesegment_results_table1.tex}

To evaluate the effectiveness of \papername{} on the primary task of lane segment detection, we compare against recent state-of-the-art methods on the OpenLane-V2 validation set.
All models are trained for 24 epochs using identical ResNet-50~\cite{he2016deep} backbones and comparable training settings to ensure a fair comparison.
We report mAP, AP$_{ls}$, AP$_{ped}$, and TOP$_{lsls}$ in Table~\ref{tab:lanseg_results}.

\papername{} achieves the best overall detection performance with an mAP of 34.5, surpassing the previous state-of-the-art Topo2Seq~\cite{yang2025topo2seq} by +0.9.
Improvements are consistent across geometric metrics, with gains of +0.2 in AP$_{ls}$ and +1.5 in AP$_{ped}$.
The improvement in AP$_{ped}$ is particularly notable, indicating stronger performance on pedestrian crossings, which are geometrically distinct and topologically sparse structures.

Notably, \papername{} achieves these improvements with a more compact model.
\papername{} contains 48M parameters, compared to 52M for Topo2Seq and 53M for TopoLogic.
This indicates that the gains are not simply a consequence of increased model capacity, but rather stem from the geometry-aware decoder design.

Compared to TopoLogic~\cite{fu2024topologic}, which achieves strong topology reasoning performance, \papername{} narrows the gap in TOP$_{lsls}$ while delivering higher detection accuracy for both lane segments and pedestrian crossings.
While Topo2Seq introduces an additional decoder branch and auxiliary training objective to improve convergence, \papername{} achieves these improvements within a unified single-decoder architecture.

Overall, these results demonstrate that \papername{} advances the state-of-the-art on lane segment detection, improving both geometric accuracy and relational consistency through geometry-aware and topology-integrated decoding.

\subsection{Qualitative Comparison}

\begin{figure*}[]
	\centering
    \includegraphics[width=\textwidth]{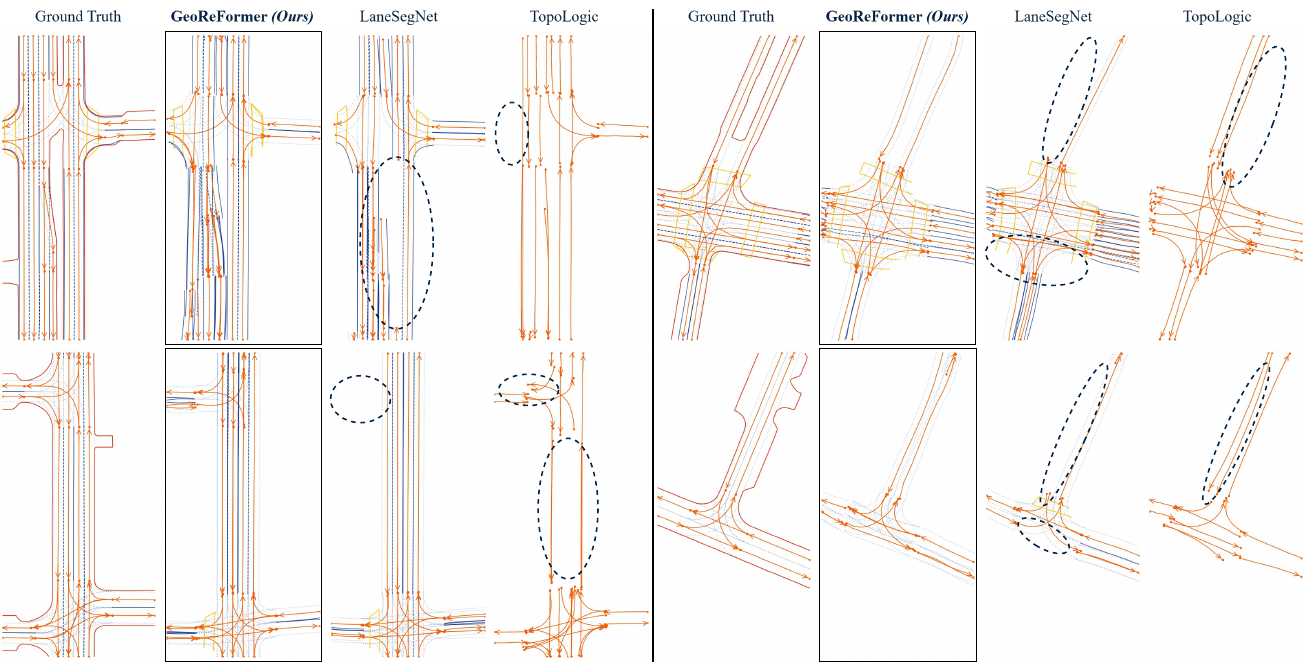}
    \caption{Qualitative comparison of lane detection on the OpenLane-V2 benchmark. From left to right: ground truth, \papername{}, LaneSegNet and TopoLogic.  The dotted ellipses show missing features correctly captured by \papername{}.}
	\label{fig:qualitative}
\end{figure*}

In the previous section, \papername{} demonstrates improved quantitative performance across multiple evaluation metrics.
We now examine representative qualitative examples to better understand how these improvements appear in predicted scene structure.
Fig.~\ref{fig:qualitative} presents side-by-side inference results from \papername{}, LaneSegNet, and TopoLogic, together with ground-truth annotations.
Lane centerlines are visualized in orange, lane boundaries in blue, and pedestrian crossings in yellow.
Topo2Seq is not included because its official implementation is not publicly available.

Across diverse scenes, \papername{} produces more complete and geometrically consistent predictions.
In comparison, LaneSegNet and TopoLogic frequently miss lane segments, intersection structures, or pedestrian crossings 
(highlighted in the dotted ellipses of Fig.~\ref{fig:qualitative})
that are correctly detected by \papername{}.
These differences are visible both near the ego vehicle and in distant regions of the BEV representation, where accurate geometric refinement becomes more challenging.

Overall, the qualitative results indicate that the quantitative improvements correspond to meaningful gains in scene reconstruction, with \papername{} producing more reliable and spatially consistent lane representations across a range of varied road layouts.

\subsection{Per-Query vs. Global Topology Gating}

\begin{figure}[]
    \centering
    \input{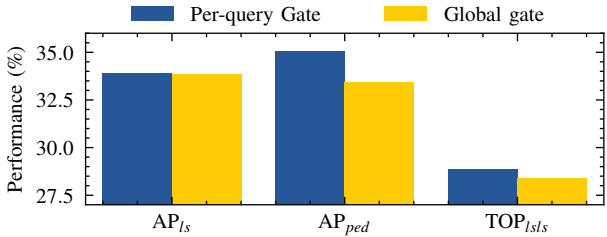}
    \caption{Comparison between per-query and global topology gating on the Lane Segment Detection task}
    \label{fig:per_query_barplot}
\end{figure}

We analyze the impact of the proposed per-query topology gate within the TopoFFN module to determine whether query-specific modulation is critical for pedestrian crossing detection.
To isolate the effect of this design choice, we compare the full \papername{} model with a variant in which the per-query gate is replaced by a single global gate shared across all queries.
All other architectural components and training settings remain identical, and both models are trained for 24 epochs under the same optimization protocol.
Results are shown in Fig.~\ref{fig:per_query_barplot}, where we report $AP_{ls}$, $AP_{ped}$, and $TOP_{lsls}$.

We observe that both designs achieve nearly identical performance on $AP_{ls}$, indicating that lane segment detection is largely unaffected by whether topology aggregation is query-specific or globally modulated.
A moderate decrease is observed in $TOP_{lsls}$, dropping from 28.9\% to 28.4\%, suggesting a slight reduction in relational consistency when topology information is applied uniformly across queries.
The most significant degradation occurs in $AP_{ped}$, which decreases from 35.0\% to 33.4\% under the global gating design.

These results indicate that indiscriminate topology aggregation can negatively impact pedestrian crossing detection, which exhibits sparse or weak connectivity in the lane graph.
The per-query gating mechanism enables selective incorporation of relational context, mitigating noise propagation to structurally isolated pedestrian crossing queries.
Overall, this experiment demonstrates that query-level topology modulation plays a critical role in improving pedestrian crossing detection within \papername{}.

\input{lsn_ablation_table.tex}

\subsection{Ablation on Geometry-Aware Decoder Components}

To isolate the contribution of the geometry-aware components in \papername{}, we conduct a controlled ablation study using the LaneSegNet baseline on the Lane Segment Detection task.
We evaluate four configurations: (i) vanilla LaneSegNet, (ii) LaneSegNet with Geometric Prior Initialization (GPI), (iii) LaneSegNet with Bounded Coordinate-Space Polyline Refinement (BCPR), and (iv) LaneSegNet with both GPI and BCPR.
All variants are trained for 12 epochs under identical settings.
We report the results in Table~\ref{tab:lanesegnet_ablations}.

We make two key observations.
First, introducing either GPI or BCPR independently degrades performance relative to the baseline.
When GPI is applied without modifying the refinement mechanism, structured priors are updated through unconstrained logit-space residuals, disrupting their geometric structure.
Conversely, applying BCPR without geometric priors constrains refinement around randomly initialized queries, limiting convergence to meaningful lane geometries.

Second, when GPI and BCPR are combined, performance improves over the baseline across geometric metrics.
The joint configuration achieves the highest $mAP$ and $AP_{ped}$ while maintaining comparable $TOP_{lsls}$ performance.
This demonstrates that structured initialization and bounded refinement are complementary: geometric priors provide semantically aligned starting points, while bounded updates preserve and progressively adjust them.

Overall, these results highlight the importance of coupling geometric prior initialization with coordinate-space refinement.
Structured priors alone are ineffective when subsequent updates are unconstrained, while bounded refinement without meaningful initialization restricts learning around arbitrary queries.
Only their combination enables consistent geometric improvement without degrading topology performance, supporting the central design principle of \papername{}.

\subsection{Sensitivity to Residual Scaling Factor $s$}

\begin{figure}[]
    \centering
    \input{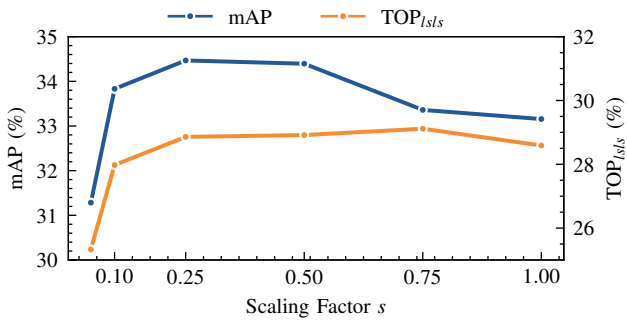}
    \caption{Sensitivity of \papername{} to the residual scaling factor $s$. 
Performance trends for mAP and TOP$_{lsls}$ across different scaling values.}
    \label{fig:s_sensitivity}
\end{figure}

We analyze the sensitivity of \papername{} to the residual scaling factor $s$ introduced in Sec.~\ref{sec:bounded_refinement}, which controls the magnitude of bounded coordinate-space updates.
To assess its effect, we train six models with $s \in \{0.05, 0.10, 0.25, 0.50, 0.75, 1.0\}$ for 24 epochs under identical settings.
Fig.~\ref{fig:s_sensitivity} shows the resulting trends, with $s$ on the x-axis and mAP and TOP$_{lsls}$ on the left and right y-axes, respectively.

We make two key observations.
First, topology performance remains largely stable once $s$ exceeds 0.1, indicating that relational reasoning is relatively insensitive to moderate variations in refinement magnitude.
Second, geometric accuracy (mAP) exhibits a pronounced sensitivity to $s$.
Very small values ($s \leq 0.1$) lead to under-refinement, preventing adequate adjustment of geometric priors.
Conversely, large values ($s \geq 0.5$) degrade performance, suggesting that overly aggressive updates destabilize local polyline structure.
The best performance is achieved within an intermediate range of $s$, supporting our hypothesis that bounded refinement introduces a stability–expressivity trade-off.
These results confirm the importance of constraining coordinate-space updates so that priors are refined within local neighborhoods rather than undergoing global shifts.

\subsection{Interpretability of Geometric Priors}

\begin{figure*}[t]
	\centering
    \includegraphics[width=\textwidth]{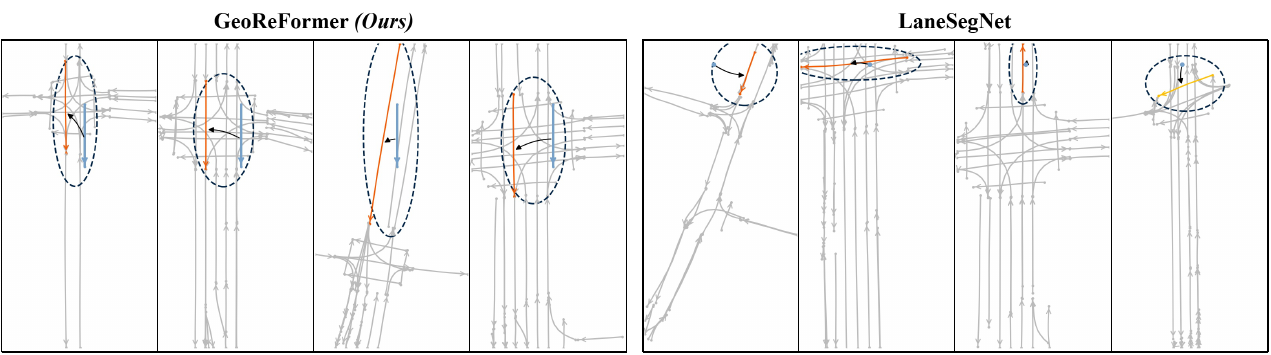}
    \caption{Interpretability comparison between \papername{} and LaneSegNet. Selected geometric priors and learned reference points are shown in blue, with corresponding detections highlighted (centerlines in orange, pedestrian crossings in yellow). \papername{} priors exhibit consistent geometric roles across scenes, while LaneSegNet queries produce heterogeneous detections. Each column corresponds to a different scene.}
	\label{fig:interpretability}
\end{figure*}

Alongside improved quantitative performance, \papername{} also improves the interpretability of query representations.
We visualize detections associated with a selected geometric prior from \papername{} across multiple scenes and compare them to detections produced by a learned reference point in LaneSegNet (Fig.~\ref{fig:interpretability}).

The geometric prior in \papername{} exhibits consistent semantic specialization.
Across scenes, its detections share similar geometric structure and orientation, repeatedly capturing lane centerlines oriented roughly opposite the ego direction.
In contrast, learned reference points in LaneSegNet do not exhibit such specialization.
A single query may produce detections with multiple lane orientations and pedestrian crossings, indicating a more entangled mapping between queries and predicted structures.

These results suggest that the structured geometric priors in \papername{} encourage queries to converge toward consistent geometric roles during training.
Because each prior encodes a geometric prototype, its associated detections form a coherent family of lane structures.
Consequently, the learned queries become human-interpretable: a geometric prior directly indicates the class of lane structures it is likely to detect, unlike models with unconstrained learned queries where query semantics are entangled.

\section{CONCLUSION \& FUTURE WORK}

We presented \papername{}, a geometry-aware transformer for joint 3D lane detection and topology reasoning.
Unlike generic DETR-style decoders, our approach embeds geometric priors, bounded coordinate-space refinement, and per-query gated topology aggregation within the decoder.
Experiments on OpenLane-V2 show that coupling geometric priors with locality-preserving refinement improves both geometric accuracy and topology consistency over strong transformer baselines, with ablations confirming that the two components are complementary, as well as qualitative results showing more reliable
and spatially consistent lane representations.
Additionally, we showed that \papername{} leads to more interpretable learned queries.
Future work includes extending geometry-aware refinement to temporal or streaming settings and learning adaptive priors for diverse road layouts.

\bibliographystyle{IEEEtran}
\bibliography{refs}

\end{document}

%% file: vars.tex
\newcommand{\papername}{GeoReFormer}

%% file: lanesegment_results_table1.tex
\begin{table}[]
\centering
\caption{Performance comparison with state-of-the-art methods on OpenLane-V2 benchmark on lane segment detection}
\label{tab:lanseg_results}
\resizebox{\columnwidth}{!}{%
\begin{tabular}{@{}c|c|cccc|c@{}}
\toprule
\textbf{Method}                      & \textbf{Venue} & \textbf{mAP}  & \textbf{AP$_{ls}$} & \textbf{AP$_{ped}$} & \textbf{TOP$_{lsls}$} & \textbf{\# Params} \\ \midrule
MapTR~\cite{liaomaptr}      & ICLR23    & 27.0 & 25.9 & 28.1 & -             & -   \\
MapTRv2~\cite{liao2025maptrv2}    & IJCV24    & 28.5 & 26.6 & 30.4 & -             & -   \\
TopoNet~\cite{li2023toponet}    & Arxiv23   & 23.0 & 23.9 & 22.0 & -             & -   \\
LaneSegNet~\cite{li2023lanesegnet} & ICLR24    & 32.6 & 32.3 & 32.9 & 25.4          & 45M \\
TopoLogic~\cite{fu2024topologic}  & NeurIPS24 & 33.2 & 33.0 & 33.4 & \textbf{30.8} & 53M \\
Topo2Seq~\cite{yang2025topo2seq}   & AAAI25    & 33.6 & 33.7 & 33.5 & 26.9          & 52M \\ \midrule
\textit{\textbf{GeoReFormer (Ours)}} & -              & \textbf{34.5} & \textbf{33.9}      & \textbf{35.0}       & 28.9                  & 48M                \\ \bottomrule
\end{tabular}%
}
\end{table}

%% file: lsn_ablation_table.tex
\begin{table}[]
\caption{Ablation analysis of Geometric Prior Initialization (GPI) and Bounded Coordinate-Space Polyline Refinement (BCPR) on LaneSegNet.}
\label{tab:lanesegnet_ablations}
\resizebox{\columnwidth}{!}{%
\begin{tabular}{@{}ccccc@{}}
\toprule
\textbf{Method}       & \textbf{mAP}  & \textbf{AP$_{ls}$} & \textbf{AP$_{ped}$} & \textbf{TOP$_{lsls}$} \\ \midrule
LaneSegNet            & 24.0          & 24.5               & 23.6                & \textbf{19.9}         \\
LaneSegNet + GPI       & 22.5          & 23.1               & 21.9                & 18.8                  \\
LaneSegNet + BCPR      & 9.7           & 9.4                & 9.9                 & 4.8                   \\
LaneSegNet + BCPR + GPI & \textbf{25.7} & \textbf{24.7}      & \textbf{26.7}       & 19.8                  \\ \bottomrule
\end{tabular}%
}
\end{table}

%% file: refs.bib
@inproceedings{bachlechner2021rezero,
  title={Rezero is all you need: Fast convergence at large depth},
  author={Bachlechner, Thomas and Majumder, Bodhisattwa Prasad and Mao, Henry and Cottrell, Gary and McAuley, Julian},
  booktitle={Uncertainty in artificial intelligence},
  pages={1352--1361},
  year={2021},
  organization={PMLR}
}

@inproceedings{he2016deep,
  title={Deep residual learning for image recognition},
  author={He, Kaiming and Zhang, Xiangyu and Ren, Shaoqing and Sun, Jian},
  booktitle={Proceedings of the IEEE conference on computer vision and pattern recognition},
  pages={770--778},
  year={2016}
}

@inproceedings{lin2017feature,
  title={Feature pyramid networks for object detection},
  author={Lin, Tsung-Yi and Doll{\'a}r, Piotr and Girshick, Ross and He, Kaiming and Hariharan, Bharath and Belongie, Serge},
  booktitle={Proceedings of the IEEE conference on computer vision and pattern recognition},
  pages={2117--2125},
  year={2017}
}

@article{li2024bevformer,
  title={Bevformer: learning bird’s-eye-view representation from lidar-camera via spatiotemporal transformers},
  author={Li, Zhiqi and Wang, Wenhai and Li, Hongyang and Xie, Enze and Sima, Chonghao and Lu, Tong and Yu, Qiao and Dai, Jifeng},
  journal={IEEE Transactions on Pattern Analysis and Machine Intelligence},
  volume={47},
  number={3},
  pages={2020--2036},
  year={2024},
  publisher={IEEE}
}

@inproceedings{wang2023openlanev2,
  title={OpenLane-V2: A Topology Reasoning Benchmark for Unified 3D HD Mapping}, 
  author={Wang, Huijie and Li, Tianyu and Li, Yang and Chen, Li and Sima, Chonghao and Liu, Zhenbo and Wang, Bangjun and Jia, Peijin and Wang, Yuting and Jiang, Shengyin and Wen, Feng and Xu, Hang and Luo, Ping and Yan, Junchi and Zhang, Wei and Li, Hongyang},
  booktitle={NeurIPS},
  year={2023}
}

@article{li2023toponet,
  title={Graph-based Topology Reasoning for Driving Scenes},
  author={Li, Tianyu and Chen, Li and Wang, Huijie and Li, Yang and Yang, Jiazhi and Geng, Xiangwei and Jiang, Shengyin and Wang, Yuting and Xu, Hang and Xu, Chunjing and Yan, Junchi and Luo, Ping and Li, Hongyang},
  journal={arXiv preprint arXiv:2304.05277},
  year={2023}
}

@inproceedings{li2023lanesegnet,
  title={LaneSegNet: Map Learning with Lane Segment Perception for Autonomous Driving},
  author={Li, Tianyu and Jia, Peijin and Wang, Bangjun and Chen, Li and Jiang, Kun and Yan, Junchi and Li, Hongyang},
  booktitle={ICLR},
  year={2024}
}

@inproceedings{fu2024topologic,
 author = {Fu, Yanping and Liao, Wenbin and Liu, Xinyuan and Xu, Hang and Ma, Yike and Zhang, Yucheng and Dai, Feng},
 booktitle = {Advances in Neural Information Processing Systems},
 pages = {61658--61676},
 title = {TopoLogic: An Interpretable  Pipeline for Lane Topology Reasoning on Driving Scenes},
 volume = {37},
 year = {2024}
}

@inproceedings{futopopoint,
  title={TopoPoint: Enhance Topology Reasoning via Endpoint Detection in Autonomous Driving},
  author={Fu, Yanping and Liu, Xinyuan and others},
  booktitle={The Thirty-ninth Annual Conference on Neural Information Processing Systems}
}

@inproceedings{lv2025t2sg,
  title={T2sg: Traffic topology scene graph for topology reasoning in autonomous driving},
  author={Lv, Changsheng and Qi, Mengshi and Liu, Liang and Ma, Huadong},
  booktitle={Proceedings of the Computer Vision and Pattern Recognition Conference},
  pages={17197--17206},
  year={2025}
}

@inproceedings{liaomaptr,
  title={MapTR: Structured Modeling and Learning for Online Vectorized HD Map Construction},
  author={Liao, Bencheng and Chen, Shaoyu and Wang, Xinggang and Cheng, Tianheng and Zhang, Qian and Liu, Wenyu and Huang, Chang},
  booktitle={The Eleventh International Conference on Learning Representations},
  year={2023}
}

@article{liao2025maptrv2,
  title={Maptrv2: An end-to-end framework for online vectorized hd map construction},
  author={Liao, Bencheng and Chen, Shaoyu and others},
  journal={International Journal of Computer Vision},
  volume={133},
  number={3},
  pages={1352--1374},
  year={2025},
  publisher={Springer}
}

@inproceedings{yang2025topo2seq,
  title={Topo2seq: Enhanced topology reasoning via topology sequence learning},
  author={Yang, Yiming and Luo, Yueru and others},
  booktitle={Proceedings of the AAAI Conference on Artificial Intelligence},
  volume={39},
  number={9},
  pages={9318--9326},
  year={2025}
}

@inproceedings{yuan2024streammapnet,
  title={Streammapnet: Streaming mapping network for vectorized online hd map construction},
  author={Yuan, Tianyuan and Liu, Yicheng and others},
  booktitle={Proceedings of the IEEE/CVF Winter Conference on Applications of Computer Vision},
  pages={7356--7365},
  year={2024}
}

@inproceedings{wang2024stream,
  title={Stream query denoising for vectorized hd-map construction},
  author={Wang, Shuo and Jia, Fan and others},
  booktitle={European Conference on Computer Vision},
  pages={203--220},
  year={2024},
  organization={Springer}
}

@article{yang2025topostreamer,
  title={Topostreamer: Temporal lane segment topology reasoning in autonomous driving},
  author={Yang, Yiming and Luo, Yueru and others},
  journal={arXiv preprint arXiv:2507.00709},
  year={2025}
}

@inproceedings{ye2025smart,
  title={Smart: Advancing scalable map priors for driving topology reasoning},
  author={Ye, Junjie and Paz, David and others},
  booktitle={2025 IEEE International Conference on Robotics and Automation (ICRA)},
  pages={3298--3304},
  year={2025},
  organization={IEEE}
}

@inproceedings{zhang2024enhancing,
  title={Enhancing online road network perception and reasoning with standard definition maps},
  author={Zhang, Hengyuan and Paz, David and others},
  booktitle={2024 IEEE/RSJ International Conference on Intelligent Robots and Systems (IROS)},
  pages={1086--1093},
  year={2024},
  organization={IEEE}
}

@inproceedings{pham2025coherent,
  title={Coherent Online Road Topology Estimation and Reasoning with Standard-Definition Maps},
  author={Pham, Khanh Son and Witte, Christian and others},
  booktitle={2025 IEEE/RSJ International Conference on Intelligent Robots and Systems (IROS)},
  pages={9886--9893},
  year={2025},
  organization={IEEE}
}

@inproceedings{carion2020end,
  title={End-to-end object detection with transformers},
  author={Carion, Nicolas and Massa, Francisco and others},
  booktitle={European conference on computer vision},
  pages={213--229},
  year={2020},
  organization={Springer}
}

@inproceedings{zhudeformable,
  title={Deformable DETR: Deformable Transformers for End-to-End Object Detection},
  author={Zhu, Xizhou and Su, Weijie and others},
  booktitle={International Conference on Learning Representations},
  year={2021}
}

@inproceedings{chen2022persformer,
  title={Persformer: 3d lane detection via perspective transformer and the openlane benchmark},
  author={Chen, Li and Sima, Chonghao and Li, Yang and others},
  booktitle={European Conference on Computer Vision},
  pages={550--567},
  year={2022},
  organization={Springer}
}

@inproceedings{liu2023vectormapnet,
  title={Vectormapnet: End-to-end vectorized hd map learning},
  author={Liu, Yicheng and Yuan, Tianyuan and Wang, Yue and Wang, Yilun and Zhao, Hang},
  booktitle={International conference on machine learning},
  pages={22352--22369},
  year={2023},
  organization={PMLR}
}

@inproceedings{ma2024roadpainter,
  title={Roadpainter: Points are ideal navigators for topology transformer},
  author={Ma, Zhongxing and Liang, Shuang and Wen, Yongkun and Lu, Weixin and Wan, Guowei},
  booktitle={European Conference on Computer Vision},
  pages={179--195},
  year={2024},
  organization={Springer}
}

@inproceedings{luo2024augmenting,
  title={Augmenting lane perception and topology understanding with standard definition navigation maps},
  author={Luo, Katie Z and Weng, Xinshuo and others},
  booktitle={2024 IEEE International Conference on Robotics and Automation (ICRA)},
  pages={4029--4035},
  year={2024},
  organization={IEEE}
}

@inproceedings{li2022hdmapnet,
  title={Hdmapnet: An online hd map construction and evaluation framework},
  author={Li, Qi and Wang, Yue and others},
  booktitle={2022 International Conference on Robotics and Automation (ICRA)},
  pages={4628--4634},
  year={2022},
  organization={IEEE}
}
